\PassOptionsToPackage{unicode}{hyperref}
\PassOptionsToPackage{hyphens}{url}
\PassOptionsToPackage{dvipsnames,svgnames,x11names}{xcolor}
\documentclass[
  12pt]{article}
\usepackage{amsthm, mathtools}
\usepackage{natbib}
\usepackage{xcolor}
\usepackage{soulpos}
\usepackage{xr-hyper}
\usepackage[colorlinks=true]{hyperref}
\usepackage[ruled,vlined,linesnumbered]{algorithm2e}
\usepackage{algpseudocode}
\usepackage{setspace}
\usepackage{soul}
\usepackage[normalem]{ulem}
\usepackage{subcaption}
\usepackage{graphicx}
\usepackage{amsmath,amssymb}
\usepackage{iftex}
\ifPDFTeX
  \usepackage[T1]{fontenc}
  \usepackage[utf8]{inputenc}
  \usepackage{textcomp} 
\else 
  \usepackage{unicode-math}
  \defaultfontfeatures{Scale=MatchLowercase}
  \defaultfontfeatures[\rmfamily]{Ligatures=TeX,Scale=1}
\fi
\usepackage{lmodern}
\ifPDFTeX\else  
\fi
\IfFileExists{upquote.sty}{\usepackage{upquote}}{}
\IfFileExists{microtype.sty}{
  \usepackage[]{microtype}
  \UseMicrotypeSet[protrusion]{basicmath} 
}{}
\makeatletter
\@ifundefined{KOMAClassName}{
  \IfFileExists{parskip.sty}{%
    \usepackage{parskip}
  }{
    \setlength{\parindent}{0pt}
    \setlength{\parskip}{6pt plus 2pt minus 1pt}}
}{
  \KOMAoptions{parskip=half}}
\makeatother
\usepackage{xcolor}
\makeatletter
\ifx\paragraph\undefined\else
  \let\oldparagraph\paragraph
  \renewcommand{\paragraph}{
    \@ifstar
      \xxxParagraphStar
      \xxxParagraphNoStar
  }
  \newcommand{\xxxParagraphStar}[1]{\oldparagraph*{#1}\mbox{}}
  \newcommand{\xxxParagraphNoStar}[1]{\oldparagraph{#1}\mbox{}}
\fi
\ifx\subparagraph\undefined\else
  \let\oldsubparagraph\subparagraph
  \renewcommand{\subparagraph}{
    \@ifstar
      \xxxSubParagraphStar
      \xxxSubParagraphNoStar
  }
  \newcommand{\xxxSubParagraphStar}[1]{\oldsubparagraph*{#1}\mbox{}}
  \newcommand{\xxxSubParagraphNoStar}[1]{\oldsubparagraph{#1}\mbox{}}
\fi
\makeatother
\usepackage{longtable,booktabs,array}
\usepackage{calc} 
\usepackage{etoolbox}
\makeatletter
\patchcmd\longtable{\par}{\if@noskipsec\mbox{}\fi\par}{}{}
\makeatother
\IfFileExists{footnotehyper.sty}{\usepackage{footnotehyper}}{\usepackage{footnote}}
\makesavenoteenv{longtable}
\usepackage{graphicx}
\makeatletter
\def\maxwidth{\ifdim\Gin@nat@width>\linewidth\linewidth\else\Gin@nat@width\fi}
\def\maxheight{\ifdim\Gin@nat@height>\textheight\textheight\else\Gin@nat@height\fi}
\makeatother
\setkeys{Gin}{width=\maxwidth,height=\maxheight,keepaspectratio}
\makeatletter
\def\fps@figure{htbp}
\makeatother

\makeatletter
\@ifpackageloaded{caption}{}{\usepackage{caption}}
\AtBeginDocument{%
\ifdefined\contentsname
  \renewcommand*\contentsname{Table of contents}
\else
  \newcommand\contentsname{Table of contents}
\fi
\ifdefined\listfigurename
  \renewcommand*\listfigurename{List of Figures}
\else
  \newcommand\listfigurename{List of Figures}
\fi
\ifdefined\listtablename
  \renewcommand*\listtablename{List of Tables}
\else
  \newcommand\listtablename{List of Tables}
\fi
\ifdefined\figurename
  \renewcommand*\figurename{Figure}
\else
  \newcommand\figurename{Figure}
\fi
\ifdefined\tablename
  \renewcommand*\tablename{Table}
\else
  \newcommand\tablename{Table}
\fi
}
\@ifpackageloaded{float}{}{\usepackage{float}}
\usepackage{placeins}

\floatstyle{ruled}
\@ifundefined{c@chapter}{\newfloat{codelisting}{h}{lop}}{\newfloat{codelisting}{h}{lop}[chapter]}
\floatname{codelisting}{Listing}

\makeatother
\makeatletter
\@ifpackageloaded{caption}{}{\usepackage{caption}}
\@ifpackageloaded{subcaption}{}{\usepackage{subcaption}}
\makeatother
\ifLuaTeX
  \usepackage{selnolig}  
\fi
\usepackage[]{natbib}
\usepackage{bookmark}
\IfFileExists{xurl.sty}{\usepackage{xurl}}{} 
\hypersetup{
  pdftitle={Temporal-Difference Calibration of Black-Box Simulators},
  pdfauthor={},
  pdfkeywords={derivative-free temporal-difference learning, few-replicate bootstrap, online inference, Polyak--Ruppert averaging, sim-to-real policy evaluation, zeroth-order stochastic approximation},
  colorlinks=true,
  linkcolor={blue},
  filecolor={blue},
  citecolor={Blue},
  urlcolor={Blue},
  pdfcreator={LaTeX via pandoc}}

\newtheorem{assumption}{Assumption}
\newtheorem{remark}{Remark}
\newtheorem{theorem}{Theorem}
\newtheorem{lemma}{Lemma}
\newtheorem{proposition}{Proposition}
\newtheorem{corollary}{Corollary}
\newcommand{\EE}{\mathbb{E}}

\newcommand{\PP}{\mathbb{P}}
\newcommand{\Var}{\mathbb{V}}

\newcommand{\CI}{\text{CI}}

\makeatletter
\renewcommand\section{\@startsection{section}{1}{\z@}%
  {-0.01ex \@plus -0.5ex \@minus -.2ex}{2.3ex \@plus.2ex}{\normalfont\Large\bfseries}}
\renewcommand\subsection{\@startsection{subsection}{2}{\z@}%
  {-0.75ex\@plus -0.5ex \@minus -.2ex}{1.5ex \@plus .2ex}{\normalfont\large\bfseries}}
\def\@afterheading{\@nobreaktrue
  \everypar{\if@nobreak\@nobreakfalse\clubpenalty\@clubpenalty
    \if@afterindent\else{\setbox\z@\lastbox}\fi
    \else\clubpenalty\@clubpenalty\everypar{}\fi}}
\makeatother
\allowdisplaybreaks[1]
\begin{document}
\def\spacingset#1{\renewcommand{\baselinestretch}%
{#1}\small\normalsize} \spacingset{1}
\spacingset{1.5} 
\makeatletter
\g@addto@macro\normalsize{%
\setlength{\abovedisplayskip}{2pt plus 2pt minus 1pt}%
\setlength{\belowdisplayskip}{2pt plus 2pt minus 1pt}%
\setlength{\abovedisplayshortskip}{0pt plus 2pt}%
\setlength{\belowdisplayshortskip}{1pt plus 2pt}}
\makeatother
\normalsize
\setlength{\jot}{0pt}
{
 \bigskip
 \bigskip
 \bigskip
 \begin{center}
	   {\Large\bf \ifdefined\textls\textls[-25]{Temporal-Difference Calibration of Black-Box Simulators}\else Temporal-Difference Calibration of Black-Box Simulators\fi}

\bigskip
{\normalsize
Min Zeng$^{1}$,
Xin Wen$^{2}$,
Xi Chen$^{2}$,
Yichen Zhang$^{3}$ and
Wen-Xin Zhou$^{4}$\par}

\vspace{0.9em}

{\small
$^{1}$College of Computing,
City University of Hong Kong\par

$^{2}$Stern School of Business,
New York University\par

$^{3}$Daniels School of Business,
Purdue University\par

$^{4}$Department of Information and Decision Science,
University of Illinois Chicago\par}



\end{center}
 \medskip
} 
\bigskip

\begin{abstract}

We develop a derivative-free temporal-difference (TD) framework for calibrating black-box simulators using one-step transitions from a target system and noisy simulator-generated continuation values. The calibration parameter is defined through Bellman orthogonality, recovering the physical parameter under correct simulator specification and yielding a near-optimal $L_2$ approximation of the policy value under misspecification. We propose three TD algorithms based on random-direction finite differences, distinguished by their query costs and reuse of noisy simulator evaluations. We establish almost-sure convergence, last-iterate mean-square error bounds, and central limit theorems for Polyak-Ruppert averages across four query-noise regimes. Our analysis reveals a specification-dependent bias phenomenon: for two algorithms, the finite-difference bias of the TD estimating equation vanishes at the calibration target under correct specification, even when the gradient surrogate remains biased. This permits a fixed perturbation radius and rate exponents approaching $1/2$ in the simulator rollout budget, whereas misspecification generally recovers the classical Kiefer-Wolfowitz rates. We also develop a few-replicate online bootstrap that yields Student-$t$ calibrated confidence intervals under negligible query noise, and identify a covariance mismatch that can invalidate conventional multiplier inference along Markov trajectories. Finally, a second-stage TD recursion corrects the remaining value discrepancy without additional simulator queries.
\end{abstract}
\noindent%
\vfill

\section{Introduction}
\label{sec:introduction}

Simulation models play an important role in policy evaluation and decision-making, particularly when direct experimentation with the target system is costly or impractical. In many sim-to-real applications, simulation remains integral to policy evaluation even after deployment \citep{coraci2025scalable}. The deployed system provides observations of states, rewards, and one-step transitions, while a simulator, initialized at the observed states and operated under the same policy, generates estimates of future returns. Such simulators often encode structural or physical knowledge that is difficult to preserve in a generic statistical predictor \citep{song2022online}. For instance, in robotic locomotion, simulator parameters may represent friction coefficients, the center of mass, and other physically interpretable quantities \citep{yu2019sim}. Calibrating these parameters to the target system is therefore essential for obtaining reliable policy evaluations while retaining the structural interpretability of the simulator.

A fundamental challenge is that the target system and the simulator provide different forms of information. Observations from the target system may be limited to one-step transitions, whereas simulator rollouts can be generated repeatedly at selected states and parameter values. Moreover, many simulators are accessible only through noisy function evaluations, without parameter gradients. Temporal-difference (TD) learning updates value estimates using one-step transitions and continuation values, without requiring complete trajectories or realized long-term returns \citep{Sutton1988}. Using continuation values from simulator rollouts, calibration can proceed online, with additional simulation improving the update precision without further observations from the target system.

We define the calibration parameter through Bellman orthogonality, requiring the simulator's one-step Bellman discrepancy to be orthogonal to its value gradient. Since the policy value is not directly observed, recovering the conventional $L_2$ calibration target \citep{tuo2015efficient, tuo2016theoretical} through Bellman residuals requires an additional instrument. TD calibration avoids estimating this instrument. We establish an oracle inequality showing that the calibrated value achieves $L_2$ approximation error within a constant factor of the best approximation in the simulator family. Under correct specification, the calibration target coincides with the physical parameter when the TD root is unique.

Implementing this calibration presents another statistical challenge because the gradient of the simulator value is unavailable. Standard TD methods require the gradient to form the update direction \citep{Sutton1988,tsitsiklis1997analysis}. Black-box simulators return only noisy value evaluations, and random-direction finite differences approximate the missing gradient \citep{KieferWolfowitz1952,Spall1992}. A smaller perturbation radius reduces finite-difference bias but amplifies query noise. Each update also multiplies a noisy TD residual by a noisy gradient surrogate. Reusing simulator evaluations across these factors introduces dependence that can affect both the bias and variance of the update. How does replacing exact gradients with noisy function evaluations affect the convergence and statistical efficiency of TD calibration? How should query designs, perturbation radii, and simulator replication be chosen under a simulation budget, and how does this choice depend on simulator accuracy? Online inference adds to this budget, since each bootstrap path requires fresh simulator queries at its own parameter and costs as many rollouts as the main path. Can valid inference be obtained with a small number of bootstrap paths? 

We address these questions by developing a unified framework for derivative-free TD (DF-TD) calibration and inference. Our main contributions are summarized as follows.

\begin{itemize}
    \item First, we develop three DF-TD algorithms and characterize how their query designs affect simulation cost and statistical efficiency. DF-Ind serves as the baseline, using separate simulator queries for the TD residual and gradient surrogate. DF-Fwd and DF-Sym reuse evaluations through one-sided and symmetric differences, at the cost of introducing dependence between these noisy factors. DF-Sym retains the second-order accuracy of DF-Ind while saving one query per update. We establish almost-sure convergence and last-iterate mean-square error bounds, together with central limit theorems for Polyak--Ruppert averages across four query-noise regimes. Multiplying a noisy TD residual by a noisy gradient surrogate introduces additional variability through products of query errors, which are asymptotically non-negligible when query noise persists but negligible when it decays. Surprisingly, DF-Sym has smaller asymptotic covariance than the baseline DF-Ind under persistent noise when the query noise has kurtosis below $5$.
    
    \item Second, we show that simulator specification changes the statistical cost of derivative-free calibration through bias cancellation in the TD estimating equation. For DF-Ind and DF-Fwd, a bias of the gradient surrogate enters this equation weighted by the Bellman discrepancy. Correct specification makes this discrepancy zero at the calibration target. These algorithms can therefore calibrate consistently without eliminating gradient bias, with a rate exponent approaching $1/2$ in the rollout budget. Under misspecification, the same algorithms generally lose this cancellation, and controlling the remaining bias reduces the attainable rate exponents to $1/3$ for DF-Ind and $1/4$ for DF-Fwd.
    
    \item Third, since each bootstrap path requires fresh simulator queries at additional rollout cost, we develop an economical online multiplier bootstrap using a fixed, small number of bootstrap paths. We establish joint asymptotic normality across bootstrap deviations, yielding valid confidence intervals without trajectory storage. Under Markovian trajectory sampling, conventional multiplier bootstrap fails to capture temporal dependence: independent multipliers disrupt the telescoping Poisson remainder and reproduce only the lag-zero covariance of the TD score rather than its long-run covariance. We illustrate how this mismatch distorts coverage under misspecification through a simple Markovian example, while showing that a correctly specified simulator restores valid inference.

    \item Finally, a second-stage TD recursion can correct the value discrepancy remaining after simulator calibration without additional simulator queries. The discrepancy itself satisfies a Bellman equation driven by the TD residuals already computed during calibration, allowing the second stage to estimate it online. We establish joint asymptotic normality of the calibration and correction estimators, enabling inference on the corrected value.
\end{itemize}

Our work connects several strands of the literature. The first is derivative-free stochastic approximation, including the classical Kiefer-Wolfowitz procedure \citep{KieferWolfowitz1952}, simultaneous perturbation methods \citep{Spall1992}, random-direction optimization \citep{NesterovSpokoiny2017}, and averaged derivative-free procedures \citep{DipponRenz1997}. Particularly relevant is \cite{ChenLaiLiZhang2024}, who develop online inference for Kiefer-Wolfowitz stochastic optimization of a strongly convex objective. Their function evaluations are constructed from a common stochastic observation, whereas our setting involves a TD semigradient whose root need not minimize an objective function, and simulator evaluations introduce independent query noise that is amplified by finite differencing. The interaction between this noise and the noisy TD residual leads to additional covariance components and makes simulation replication an explicit design variable. The Bellman orthogonality structure also produces a specification-dependent finite-difference bias that is absent from conventional analyses.

A second related literature concerns calibration of imperfect computer models. \cite{KennedyOHagan2001} introduced a discrepancy-based framework, while \cite{tuo2015efficient, tuo2016theoretical} developed frequentist calibration procedures based on $L_2$ projection. Subsequent contributions include orthogonality-constrained calibration \citep{Plumlee2017}, prediction-oriented calibration with finite-sample guarantees \citep{DaiChien2018}, and functional calibration using reproducing kernel Hilbert spaces \citep{TuoEtAl2023}. In our setting, the target value function is not directly observed. We therefore calibrate the simulator using one-step transitions from the target system through Bellman orthogonality.

Our analysis also contributes to the literature on statistical inference for stochastic approximation and reinforcement learning. Online inference for stochastic gradient methods has been studied by \cite{ChenLeeTongZhang2020}, \cite{LiLouSchmidtHieberWu2025}, and \cite{HanXu2026}, among others. Random-scaling methods construct online confidence intervals from the iterate path without resampling, as in e.g., \cite{lee2022randomscaling}, 
\cite{li2026streamsgd}, and \cite{chen2025sgmm}. Related developments in online decision-making include \cite{LuoSunLiu2024}, \cite{HanSunZhang2025}, \cite{DuanLiXia2026}, and \cite{WenSunZhang2026}. For policy evaluation, \cite{ShiZhangLuSong2022ValueInferenceRL,BianEtAl2025,Shi2025Survey} study statistical inference for value functions. Work on TD-based procedures addresses estimation and inference \citep{WuLiWeiRinaldo2024,SamsonovEtAl2024,LiuTuChenZhang2025} and variance reduction through experience replay \citep{HanDaiZhu2026}. Online multiplier bootstrap for linear TD learning has been studied by \cite{RamprasadEtAl2023}. We show that this bootstrap can fail under Markov dependence. In addition, for derivative-free TD calibration, each bootstrap path requires fresh simulator queries at its own parameter, adding to the simulation cost. We therefore develop inference with a fixed, small number of paths using the few-replicate bootstrap principle of \cite{LamWang2023}.

The remainder of the paper is organized as follows. Section~\ref{sec:background} introduces the TD calibration target and establishes its approximation properties. Section~\ref{sec:methodology} presents the three derivative-free TD algorithms. Section~\ref{sec:theory} develops their  asymptotic theory and investigates the effects of query design, simulator specification, and rollout budgets. Section~\ref{sec:online_bootstrap} develops the few-replicate online bootstrap and examines its validity under Markovian sampling. Section~\ref{sec:discrepancy_correction} introduces the second-stage value discrepancy correction. Section~\ref{sec:discussion} discusses potential extensions.

\section{Temporal-Difference Calibration}
\label{sec:background}

Consider a discounted Markov decision process for the target system, \(\mathcal M=(\mathcal S,\mathcal A,\mathcal P,r,\gamma)\), with state space \(\mathcal S\), action space \(\mathcal A\), transition kernel \(\mathcal P\), bounded reward function \(r\), and discount factor \(\gamma\in[0,1)\). We evaluate a fixed stationary policy \(\pi\). The policy induces the transition kernel \(P^\pi(\cdot\mid s):= \EE_{a\sim\pi(\cdot\mid s)}[\mathcal P(\cdot\mid s,a)]\) and the mean reward \(r^\pi(s):=\EE_{a\sim\pi(\cdot\mid s)}[r(s,a)]\). Let \(\mu\) be a stationary distribution of \(P^\pi\), and write \((P^\pi f)(s):=\int f(s')\,P^\pi(ds'\mid s)\) for a function \(f\). The policy value 
\[
V^\pi(s):=\EE\bigg[\sum_{t=0}^{\infty}\gamma^t r^\pi(s_t)\,\Big|\,s_0=s\bigg],
\] is
the unique fixed point of the Bellman operator \((\mathcal T^\pi V)(s)=r^\pi(s)+
\gamma\EE_{s'\sim P^\pi(\cdot\mid s)}[V(s')]\). 

For \(\theta\in\mathbb R^d\), the simulator is reset at state \(s\) and run under the same policy \(\pi\). Its prediction of \(V^\pi(s)\) is the expected return
\[
V(s;\theta):=\EE_{\theta}^{\rm sim}\bigg[\sum_{k=0}^{\infty}\gamma^k r_k^{\rm sim}\,\Big|\,s_0=s,\pi\bigg].
\]
These simulator value functions form \(\mathcal V=\{V(\cdot;\theta):\theta\in\mathbb R^d\}\) with linear approximation \(V(s;\theta)=\phi(s)^\top\theta\) as a special case. We use this family to approximate \(V^\pi\), even when \(V^\pi\notin\mathcal V\).

We write \(L_2(\mu)\) for the collection of square-integrable functions, with $ \|f\|_\mu^2 := \int f(s)^2\,\mu(ds)$.
The \(L_2\) calibration target of \citet{tuo2016theoretical} is \(\theta_{L_2}\in\arg\min_\theta\|V^\pi-V(\cdot;\theta)\|_\mu\), with minimum approximation error \(\varepsilon_{\min}:=\|V^\pi-V(\cdot;\theta_{L_2})\|_\mu\). The minimizer \(\theta_{L_2}\) satisfies
\[
\EE[\{V^\pi(s)-V(s;\theta_{L_2})\}\nabla_\theta V(s;\theta_{L_2})]=0.
\]
This criterion uses the policy value \(V^\pi\), which is not directly observed in our setting.

For a one-step transition \((s,r,s')\), define the \emph{TD residual} and the \emph{Bellman discrepancy}
\[
\delta(\theta;s,r,s') := r + \gamma V(s';\theta)-V(s;\theta),~~ 
\varepsilon_{\rm B}(s;\theta) := \EE\big[ \delta(\theta;s,r,s')\mid s\big]
= (\mathcal T^\pi V(\cdot;\theta))(s)-V(s;\theta),
\]
where the conditional expectation is under the one-step law of the target system induced by \(\pi\).

The TD criterion replaces the value error in the \(L_2\) orthogonality condition by the one-step Bellman discrepancy. This gives the population semigradient operator
\[
G(\theta) :=
\EE\bigl[
\delta(\theta;s,r,s')\,
\nabla_\theta V(s;\theta)
\bigr]
=
\EE\bigl[
\varepsilon_{\rm B}(s;\theta)\,
\nabla_\theta V(s;\theta)
\bigr],
\]
where the expectation is under the stationary one-step law. We define the \emph{temporal-difference calibration target} \(\theta^*\) as a solution to \(G(\theta^*)=0\). Assumption~\ref{ass:regularity}(1) below assumes that a root exists and imposes a global stability condition that implies its uniqueness. Because the estimating equation uses transitions under \(\pi\), weighted by \(\mu\), the target generally depends on $(\pi, \gamma, \mu)$, and calibrates the simulator for predicting the value of \(\pi\).

The relation between value error and Bellman discrepancy determines how the TD target differs from the \(L_2\) target. Since \(V^\pi\) is the fixed point of \(\mathcal T^\pi\), for any \(\theta\) such that \(V(\cdot;\theta)\in L_2(\mu)\),
\begin{equation}
\label{eq:value_discrepancy_resolvent}
V^\pi-V(\cdot;\theta)
=
(I-\gamma P^\pi)^{-1}\varepsilon_{\rm B}(\cdot;\theta).
\end{equation}

Both targets can therefore be expressed through the instrument equation \(\EE[W(s)\varepsilon_{\rm B}(s;\theta)]=0\) \citep{ChenQi2022,DuanWangWainwright2024}. TD calibration uses \(W(s)=\nabla_\theta V(s;\theta)\), whereas the \(L_2\) first-order condition uses \(W=(I-\gamma(P^\pi)^*)^{-1}\nabla_\theta V\), where \((P^\pi)^*\) denotes the \(L_2(\mu)\)-adjoint of \(P^\pi\). 
Proposition S1 in the Appendix establishes this 
correspondence. Accordingly, the \(L_2\) calibration instrument involves a discounted accumulation of value-function gradients over current and past states, which is not directly available from one-step observations.

We call the simulator family \emph{correctly specified}
(or \emph{perfect}, following \citet{tuo2015efficient}) if
\(V(\cdot;\theta^\dagger)=V^\pi(\cdot)\)
\(\mu\)-almost everywhere for some \(\theta^\dagger\),
equivalently \(\varepsilon_{\rm B}(\cdot;\theta^\dagger)=0\)
\(\mu\)-almost everywhere, and \emph{misspecified}
(or \emph{imperfect}) otherwise. We call \(\theta^\dagger\)
a value-matching parameter. Correct specification refers
to the value function under the given policy and discount
factor, and does not require the simulator's transition
and reward mechanisms to match those of the target system.
For a correctly specified simulator, \(\theta^\dagger\)
satisfies the estimating equation for every admissible
instrument, whereas under misspecification, the roots
generally depend on the instrument. The next proposition
shows when the TD and \(L_2\) targets coincide and bounds
their difference under misspecification.

\begin{proposition}[Temporal-difference calibration target]
\label{prop:bellman_calibration_target}
\textit{(i) Correctly specified simulator.} If the simulator is correctly specified with parameter \(\theta^\dagger\) and \(\nabla_\theta V(\cdot;\theta^\dagger)\) is square-integrable, then \(G(\theta^\dagger)=0\). If, in addition, the TD root is unique, then \(\theta^\dagger=\theta^*\), and \(\theta^*=\theta_{L_2}\) whenever the \(L_2\) projection is unique. \textit{(ii) Error bound for the value function.} For any \(\theta\) with \(V(\cdot;\theta)\in L_2(\mu)\), \(\|V^\pi-V(\cdot;\theta)\|_\mu \le(1-\gamma)^{-1}\|\varepsilon_{\rm B}(\cdot;\theta)\|_\mu\). \textit{(iii) Oracle inequality.} Suppose Assumption~\ref{ass:regularity} in Section~\ref{sec:theory} holds and an \(L_2\) projection \(\theta_{L_2}\) exists. Then
\[
\|\theta^*-\theta_{L_2}\|
\le
\frac{\gamma\,\|\mathcal E_1\|_\mu}{L}\,\varepsilon_{\min} ~~\mbox{ and }~~
\bigl\|V^\pi-V(\cdot;\theta^*)\bigr\|_\mu
\le
\Bigl(1+\frac{\gamma\,\|\mathcal E_1\|_\mu^2}{L}\Bigr)\varepsilon_{\min},
\]
where \(L\) and \(\mathcal E_1\) are the stability constant and the gradient envelope in Assumption~\ref{ass:regularity}, respectively.
\end{proposition}
Under correct specification, the TD  and \(L_2\) target both equal the value-matching parameter $\theta^\dagger$, provided they are unique.
Under misspecification, the targets may differ, but Part~(iii) shows that the TD-calibrated value achieves an \(L_2\) error within a constant factor of the best approximation in the simulator family. 
Inference concerns \(\theta^*\) and the calibrated value \(V(s;\theta^*)\) at prespecified reporting states. Section~\ref{sec:discrepancy_correction} estimates the remaining discrepancy \(V^\pi-V(\cdot;\theta^*)\).

The Bellman discrepancy also affects the local stability of the TD root. Let \(\Sigma_\nabla:=\EE[\nabla_\theta V(s;\theta^*) \nabla_\theta V(s;\theta^*)^\top]\) and, when \(G\) is differentiable at \(\theta^*\), write \(J^*:=\nabla G(\theta^*)\).

\begin{proposition}[Local stability and first-order identification]
\label{prop:stability_identifiability}
Under Assumption~\ref{ass:regularity}(2), \(G\) is continuously differentiable at \(\theta^*\) with
\[
J^*
=
\EE\Bigl[
\nabla_\theta V(s;\theta^*)
\bigl\{\gamma\nabla_\theta V(s';\theta^*)
-\nabla_\theta V(s;\theta^*)\bigr\}^\top
\Bigr]
+
\EE\bigl[
\varepsilon_{\rm B}(s;\theta^*)\,
\nabla_\theta^2V(s;\theta^*)
\bigr].
\]
Moreover, \((J^* + J^{*\top})/2\preceq-(1-\gamma)\Sigma_\nabla +\|\varepsilon_{\rm B}(\cdot;\theta^*)\|_\mu\|\mathcal E_2\|_\mu I_d\). In particular, \(J^*\) is Hurwitz whenever \(\|\varepsilon_{\rm B}(\cdot;\theta^*)\|_\mu\|\mathcal E_2\|_\mu <(1-\gamma)\lambda_{\min}(\Sigma_\nabla)\). At a correctly specified root, $J^*$ is Hurwitz if and only if \(\Sigma_\nabla\succ0\).
\end{proposition}

Under correct specification, \(\Sigma_\nabla\succ0\) characterizes local identification through the sensitivities of the value function and is also exactly the condition for local stability of the TD recursion. Under misspecification, the additional term in \(J^*\) couples the Bellman discrepancy with simulator curvature and can compromise stability even when \(\Sigma_\nabla\succ0\).

We first consider stationary i.i.d.\ one-step sampling \citep{SamsonovEtAl2024,WuLiWeiRinaldo2024}. Extensions to uniformly geometrically ergodic Markov trajectories are provided in Appendix.

\begin{assumption}[i.i.d.\ stationary one-step sampling]
\label{as:iid_sampling}
The tuples \(\{(s_t,a_t,r_t,s_t')\}_{t\ge1}\) are i.i.d. draws from the stationary one-step distribution induced by $\pi$, where \(s_t\sim\mu\), \(a_t\sim\pi(\cdot\mid s_t)\), \(r_t=r(s_t,a_t)\), and \(s_t'\sim\mathcal P(\cdot\mid s_t,a_t)\).
\end{assumption}

Under Assumption~\ref{as:iid_sampling}, the benchmark TD algorithm with access to exact gradients follows the semigradient recursion
\[
\theta_t
=
\theta_{t-1}
+
\eta_t\,
\delta(\theta_{t-1};s_t,r_t,s_t')\, \nabla_\theta V(s_t;\theta_{t-1}),
\]
where \(\{\eta_t\}_{t\ge1}\) denotes the step-size sequence. The conditional mean increment is \(\eta_t G(\theta_{t-1})\).

Under standard stochastic-approximation conditions, the Polyak--Ruppert averaged estimator \(\bar{\theta}_T:=T^{-1}\sum_{t=1}^T\theta_t\) satisfies \citep{PolyakJuditsky1992}
\begin{equation}
\label{eq:pj_clt_fo}
\sqrt{T}\,
(\bar{\theta}_T-\theta^*)
\Longrightarrow
\mathcal N\bigl(
0,\;
(-J^*)^{-1}
\Sigma_{\rm TD}
(-J^*)^{-{\top}}
\bigr),
\end{equation}
where \( \Sigma_{\rm TD} := \EE[ (\delta_t^*)^2\, \nabla_\theta V(s_t;\theta^*)\, \nabla_\theta V(s_t;\theta^*)^\top ], \) with \(\delta_t^*:=\delta(\theta^*;s_t,r_t,s_t')\).

\begin{remark}[Bellman discrepancy and TD variance]
\label{rem:td_variance_decomposition}
Since \(\EE[\delta_t^*\mid s_t]=\varepsilon_{\rm B}(s_t;\theta^*)\), the TD score covariance with exact gradients can be decomposed as
\[
\Sigma_{\rm TD}
=
\EE\bigl[
\Var(\delta_t^*\mid s_t)\,
\nabla_\theta V(s_t;\theta^*)
\nabla_\theta V(s_t;\theta^*)^\top
\bigr]
+
\EE\bigl[
\varepsilon_{\rm B}(s_t;\theta^*)^2\,
\nabla_\theta V(s_t;\theta^*)
\nabla_\theta V(s_t;\theta^*)^\top
\bigr].
\]
The first term captures variability from one-step transitions, while the second reflects the contribution of the state-dependent Bellman discrepancy. The latter vanishes when the simulator is correctly specified and the TD root is unique.
\end{remark}

\section{Methodology}
\label{sec:methodology}

The TD update requires gradients that are unavailable under black-box simulator access. We approximate these gradients using random-direction finite differences of noisy simulator evaluations and combine them with noisy TD residuals to construct derivative-free updates.

\subsection{Noisy value queries}
\label{subsec:noisy_queries}

In the intended application, a value query is produced by resetting the
black-box simulator at a state \(s\) of the target system, setting its parameter to
\(\theta\), and rolling the fixed policy forward. One rollout returns a random
discounted cumulative reward. Averaging \(M_t\) independent rollouts produces
one noisy evaluation of \(V(s;\theta)\), with higher precision as \(M_t\)
increases. We use the following benchmark model for the returned evaluation.
\begin{assumption}
\label{as:noisy}
At iteration \(t\), each fresh value query at a possibly data-dependent
state--parameter location \((s,\theta)\) returns
\[
\tilde V_{j,t}(s;\theta)
=
V(s;\theta)+\epsilon_{j,t},
\]
where the standardized noises
\(\{\epsilon_{j,t}/\sigma_t\}_{j,t}\) have a common centered, symmetric
distribution with unit variance and a finite fourth moment. They are independent
across fresh calls and all other sources of randomness. The
deterministic sequence \(\{\sigma_t^2\}\) is positive and non-increasing.
\end{assumption}
We refer to \(\epsilon_{j,t}\) as query noise and to \(\sigma_t^2\) as
its variance scale. Assumption~\ref{as:noisy} takes this variance to be
independent of the state and parameter. 
Appendix considers
location-dependent variance. For independent rollouts with a common return
variance, averaging \(M_t\) rollouts gives query variance proportional to
\(M_t^{-1}\). 
Let \(L_{\rm alg}\) be the number of fresh value queries per update and let \(C_T^{\rm main}\) denote the number of simulator rollouts used by the main path through iteration \(T\). When each returned value averages
\(M_t\) rollouts,
\begin{equation}
\label{eq:main_rollout_budget}
C_T^{\rm main}=L_{\rm alg}\sum_{t=1}^T M_t.
\end{equation}

\subsection{Random-direction surrogates and finite-difference bias}
\label{subs:finite-difference_bias}
To separate finite-difference bias from query noise, we first construct the
gradient surrogates using noise-free value evaluations.
At iteration \(t\), let \(u_t\in\mathbb R^d\) have distribution
\(\mathcal P_u\) with \(\EE[u_t]=0\) and \(\EE[u_tu_t^\top]=I_d\),
independently of the past and the current one-step sample. Let
\(\nu_t>0\) be the perturbation radius. The superscript \({\rm ex}\) denotes
this exact-query construction. The one- and two-sided surrogates are
\begin{align}
\hat{\nabla}^{\rm ex}V(s_t;\theta_{t-1})
&=
\frac{V(s_t;\theta_{t-1}+\nu_tu_t)-V(s_t;\theta_{t-1})}{\nu_t}\,u_t,
\label{eq:zo_grad_one_sided}\\
\hat{\nabla}^{\rm ex}V(s_t;\theta_{t-1})
&=
\frac{V(s_t;\theta_{t-1}+\nu_tu_t)-V(s_t;\theta_{t-1}-\nu_tu_t)}{2\nu_t}\,u_t
\label{eq:zo_grad_two_sided}
\end{align}
The next lemma bounds the finite-difference bias of these surrogates.
\begin{lemma}[Finite-difference bias of gradient surrogate]
\label{lem:gradientbias}
Suppose the smoothness and the derivative bounds for \(j=2,3\)
in Assumption~\ref{ass:regularity}(2) hold, and
\(\EE[u_tu_t^\top]=I_d\) with \(\EE\|u_t\|^4<\infty\). Then there exist finite constants \(C_1,C_2>0\)
such that, for \(\mu\)-a.e.\ \(s\) and every \(\theta\in\mathbb R^d\),
\begin{align*}
    &\Bigl\|
\EE_{u_t}\bigl[\hat{\nabla}^{\rm ex}V(s;\theta)\bigr]
-
\nabla_\theta V(s;\theta)
\Bigr\|
\le
C_1\mathcal E_2(s)\nu_t, \quad\text{for the one-sided estimator \eqref{eq:zo_grad_one_sided}, and}\\
&\Bigl\|
\EE_{u_t}\bigl[\hat{\nabla}^{\rm ex}V(s;\theta)\bigr]
-
\nabla_\theta V(s;\theta)
\Bigr\|
\le
C_2\mathcal E_3(s)\nu_t^2, \quad\text{for the two-sided estimator \eqref{eq:zo_grad_two_sided}.}
\end{align*}
\end{lemma}
Under these general bias bounds, a two-sided difference permits a more slowly shrinking radius for the same approximation accuracy. With noisy queries, the larger radius reduces the amplification of query noise.

\subsection{Derivative-free temporal-difference learning}
\label{subs:DFTD}
We combine the gradient surrogates with noisy TD residuals to obtain three
derivative-free TD algorithms. The algorithms differ in their finite-difference
construction and in whether the residual and gradient surrogate share
simulator queries, but take the common form
\[
\theta_t=\theta_{t-1}+\eta_t\hat\delta_t\,\hat\nabla_t,
\]
where \(\hat\delta_t\) is a noisy estimate of the TD residual and
\(\hat\nabla_t\) is a noisy gradient surrogate for
\(\nabla_\theta V(s_t;\theta_{t-1})\).
The bias bounds in Lemma~\ref{lem:gradientbias} concern the surrogate alone.
For convergence, we must also control the finite-difference bias in the
complete TD increment, which depends on how the two factors are combined.

Given the one-step transition \((s_t,r_t,s_t')\) at iteration
\(t\), let \(\epsilon_{1,t},\ldots,\epsilon_{4,t}\) denote independent query noises at iteration \(t\). Define
\[
\begin{aligned}
\tilde V_{1,t}
&:=V(s_t';\theta_{t-1})+\epsilon_{1,t},
&
\tilde V_{2,t}
&:=V(s_t;\theta_{t-1})+\epsilon_{2,t},\\
\tilde V_{3,t}
&:=V(s_t;\theta_{t-1}+\nu_tu_t)+\epsilon_{3,t},
&
\tilde V_{4,t}
&:=V(s_t;\theta_{t-1}-\nu_tu_t)+\epsilon_{4,t}.
\end{aligned}
\]
\medskip
\noindent\textit{DF-Ind}  uses separate returned values in the TD residual and gradient
surrogate:
\begin{equation}
\label{eq:zo_ind_update}
\hat\delta_t
=
r_t+\gamma\tilde V_{1,t}-\tilde V_{2,t},
\qquad
\hat\nabla_t
=
\frac{\tilde V_{3,t}-\tilde V_{4,t}}{2\nu_t}\,u_t.
\end{equation}
The two factors use independent query noises but share the same one-step transition.

\medskip
\noindent\textit{DF-Fwd}
reuses the central returned value in a one-sided difference:
\begin{equation}
\label{eq:zo_fwd_update}
\hat\delta_t
=
r_t+\gamma\tilde V_{1,t}-\tilde V_{2,t},
\qquad
\hat\nabla_t
=
\frac{\tilde V_{3,t}-\tilde V_{2,t}}{\nu_t}\,u_t.
\end{equation}
Reusing \(\tilde V_{2,t}\) reduces the update to three fresh queries.
Its query noise enters both factors, coupling the TD residual and gradient
surrogate. The one-sided construction has generic first-order
finite-difference bias. For a centrally symmetric direction law, however,
the direction averages of the one- and two-sided surrogates coincide at
every radius, since \(\EE[V(s;\theta-\nu u_t)u_t]=-\EE[V(s;\theta+\nu u_t)u_t]\)
and \(\EE[u_t]=0\), so DF-Fwd and DF-Ind have the same exact-query
mean field and differ only in query cost and in the variance of the query
noise.

\medskip
\noindent\textit{DF-Sym}
uses the same perturbed pair to form a two-sided difference and the midpoint value at the current state. This removes the separate central query while retaining
second-order finite-difference accuracy:
\begin{equation}
\label{eq:zo_sym_update}
\hat\delta_t
=
r_t+\gamma\tilde V_{1,t}
-\frac{\tilde V_{3,t}+\tilde V_{4,t}}{2},
\qquad
\hat\nabla_t
=
\frac{\tilde V_{3,t}-\tilde V_{4,t}}{2\nu_t}\,u_t.
\end{equation}
We write \(O(\nu_t^{\kappa_{\rm alg}})\) for the finite-difference bias
bound of the TD increment used below, where
\(\kappa_{\rm Ind}=\kappa_{\rm Sym}=2\) and \(\kappa_{\rm Fwd}=1\).\footnote{Here \(\kappa_{\rm Fwd}=1\)
is the bound in general and used in the main text. When \(\EE[u_t^{\otimes3}]=0\),
\(\kappa_{\rm Fwd}=2\).}
Query reuse leaves these bias orders unchanged but affects the limiting
covariance under persistent query noise.

The query counts per update in \eqref{eq:main_rollout_budget} are
\(L_{\rm Ind}=4\) and \(L_{\rm Fwd}=L_{\rm Sym}=3\).
For DF-Sym, the query saved relative to DF-Ind can support more rollouts
per query under a fixed budget per update.
Algorithm~\ref{alg:zo_td} summarizes the recursions.
\begingroup
\setstretch{1.0}
\begin{algorithm}[tbp]
\caption{Derivative-free TD learning (DF-TD)}
\label{alg:zo_td}
\small
\SetAlgoLined
\DontPrintSemicolon
\KwIn{$\theta_0$, step-sizes $\{\eta_t\}$, radius $\{\nu_t\}$, horizon $T$, variant in $\{\mathrm{DF\mbox{-}Ind},\mathrm{DF\mbox{-}Fwd},\mathrm{DF\mbox{-}Sym}\}$}
\KwOut{$\{\theta_t\}_{t=1}^T$, $\{\bar\theta_t\}_{t=1}^T$}
Initialize $\bar\theta_0\gets\theta_0$\;
\For{$t=1,\dots,T$}{
    Sample $(s_t,r_t,s_t')$ and $u_t\sim\mathcal P_u$\;
    \uIf{\textbf{variant} = DF-Ind}{
        Query $\tilde V_{1,t}, \tilde V_{2,t}, \tilde V_{3,t}, \tilde V_{4,t}$ and
        compute $(\hat\delta_t,\hat\nabla_t)$ via \eqref{eq:zo_ind_update}\;
    }
    \uElseIf{\textbf{variant} = DF-Fwd}{
        Query $\tilde V_{1,t}, \tilde V_{2,t}, \tilde V_{3,t}$ and
        compute $(\hat\delta_t,\hat\nabla_t)$ via \eqref{eq:zo_fwd_update}\;
    }
    \uElseIf{\textbf{variant} = DF-Sym}{
        Query $\tilde V_{1,t}, \tilde V_{3,t}, \tilde V_{4,t}$ and
        compute $(\hat\delta_t,\hat\nabla_t)$ via \eqref{eq:zo_sym_update}\;
    }
    
    $\theta_t\gets \theta_{t-1}+\eta_t(\hat\delta_t\hat\nabla_t)$\;
    $\bar\theta_t\gets \frac{t-1}{t}\bar\theta_{t-1}+\frac{1}{t}\theta_t$\;
}
\end{algorithm}
\endgroup

\section{Theoretical Analysis}
\label{sec:theory}

We analyze how finite-difference bias and amplified query noise affect
the DF-TD recursion. We first state the technical assumptions and give
a unified error decomposition.
\begin{assumption}
\label{ass:regularity}
\begin{enumerate}
    \item[(1)]
    The equation \(G(\theta)=0\) has a solution \(\theta^*\). Moreover,
    there exists \(L>0\) such that, for all \(\theta\in\mathbb R^d\),
    $(\theta-\theta^*)^\top G(\theta)\le -L\|\theta-\theta^*\|^2.$
    \item[(2)] For \(\mu\)-a.e.\ \(s\), the map
    \(\theta\mapsto V(s;\theta)\) is three times continuously differentiable.
    Moreover, there exist measurable functions
    \(\mathcal E_0,\ldots,\mathcal E_3:\mathcal S\to[0,\infty)\) such that, for all
    \(\theta\in\mathbb R^d\) and \(\mu\)-a.e.\ \(s\),
\(
|V(s;\theta)-V(s;\theta^*)|\le \mathcal E_0(s)\|\theta-\theta^*\|
\), 
\(
\|\nabla_\theta^j V(s;\theta)\|\le \mathcal E_j(s),\quad j=1,\ldots,3,
\)
and
\(
\EE\Bigl[
(\delta_t^*)^2\bigl(1+\mathcal E_1(s_t)^2\bigr)+
\sum_{j=0}^1\mathcal E_j(s_t)^4+\sum_{j=2}^3\mathcal E_j(s_t)^2
+\mathcal E_1(s_t)^2\mathcal E_2(s_t)^2
\Bigr]<\infty.
\)
\end{enumerate}
\end{assumption}
Assumption~\ref{ass:regularity}(1) is a global stability condition for the mean field \(G\). Assumption~\ref{ass:regularity}(2) supplies the smoothness and moment bounds used to control the TD residual, gradient surrogates, and query-noise terms. For the Jacobian \(J^*\) introduced above, Assumption~\ref{ass:regularity}(1) implies $\frac{J^*+(J^*)^\top}{2}\preceq -L I_d$. Thus \(J^*\) is Hurwitz and nonsingular.
\begin{remark}
\label{rem:L_linear}
For the linear value family \(V(s;\theta)=\phi(s)^\top\theta\), define
$A:=\EE\bigl[\phi(s_t)(\phi(s_t)-\gamma\phi(s_t'))^\top\bigr]$ and $
b:=\EE[r_t\,\phi(s_t)].$
Then \(-G(\theta)=A\theta-b=A(\theta-\theta^*)\), and hence  Assumption~\ref{ass:regularity}(1) holds whenever the symmetric part of \(A\) is positive definite, in which case one may take
$L=\lambda_{\min}\bigl((A+A^\top)/2\bigr)
\ge(1-\gamma)\lambda_{\min}\bigl(\EE[\phi(s_t)\phi(s_t)^\top]\bigr)$.
\end{remark}

\begin{assumption}[Algorithm tuning]
\label{ass:algo_config}
\hfill
\begin{enumerate}
    \item[(1)]
    The tuning sequences follow the polynomial schedules
    \begin{equation}
    \eta_t = c_\eta t^{-\alpha},\qquad
    \nu_t = c_\nu t^{-\beta},\qquad
    \sigma_t^2 = c_\sigma t^{-\zeta},
    \label{eq:polyrate}
    \end{equation}
    where \(c_\eta,c_\nu,c_\sigma>0\),
    \(\alpha\in(1/2,1]\), \(\beta>0\), and \(\zeta\ge0\).
    \item[(2)] 
    The direction sequence $\{u_t\}_{t\ge 1}$ is i.i.d., satisfies
    $\EE[u_t]=0$, $\EE[u_tu_t^\top]=I_d$, and
    $\EE[\|u_t\|^8]<\infty$, and is independent of the stationary
    one-step sample array in Assumption~\ref{as:iid_sampling} and of the query noises on the main path and on all bootstrap paths.
\end{enumerate}
\end{assumption}
The relative decay rates of these schedules determine the balance between finite-difference bias and amplified query
noise. 
Let \(\mathcal F_t\) be the filtration of the main path, generated by the one-step samples, random directions, and query noises of the main path
through iteration \(t\), so that \(\theta_{t-1}\) is
\(\mathcal F_{t-1}\)-measurable. Write \(\hat g_t:=\hat\delta_t\hat\nabla_t\),
so that \(\theta_t=\theta_{t-1}+\eta_t\hat g_t\), and let
\(\delta_t^{\rm ex}\) and \(\hat\nabla_t^{\rm ex}\) be the exact-query TD
residual and gradient surrogate, computed from the same
\((s_t,r_t,s_t')\) and \(u_t\) but with all query noises set to zero.
Let \(N_{\delta,t}:=\hat\delta_t-\delta_t^{\rm ex}\) and
\(N_{\nabla,t}:=\hat\nabla_t-\hat\nabla_t^{\rm ex}\). For DF-Fwd, these are
\(N_{\delta,t}=\gamma\epsilon_{1,t}-\epsilon_{2,t}\) and
\(N_{\nabla,t}=(\epsilon_{3,t}-\epsilon_{2,t})u_t/\nu_t\). The
increment then splits as \(\hat g_t=g_t^{\rm ex}+q_t\), where
\(g_t^{\rm ex}:=\delta_t^{\rm ex}\hat\nabla_t^{\rm ex}\) and
\(q_t:=\delta_t^{\rm ex}N_{\nabla,t}+N_{\delta,t}\hat\nabla_t^{\rm ex}
+N_{\delta,t}N_{\nabla,t}\). Centering \(g_t^{\rm ex}\) at its conditional
mean gives the unified decomposition
\begin{equation}
\label{eq:main_decomp_poly}
\hat g_t
=
G(\theta_{t-1})+b_t^{\rm alg}+m_t+q_t,
\end{equation}
where
\(b_t^{\rm alg}:=\EE[g_t^{\rm ex}\mid\mathcal F_{t-1}]-G(\theta_{t-1})\) and
\(m_t:=g_t^{\rm ex}-\EE[g_t^{\rm ex}\mid\mathcal F_{t-1}]\).
Here \(G(\theta_{t-1})\) drives the stochastic-approximation dynamics,
\(b_t^{\rm alg}\) is the algorithm-specific finite-difference bias, \(m_t\)
is the intrinsic TD fluctuation, and \(q_t\) is the perturbation from query noise, in which the noise of the gradient queries is amplified by
\(1/\nu_t\).

Under Assumptions~\ref{as:iid_sampling}--\ref{ass:algo_config}, both \(m_t\) and \(q_t\) have conditional mean zero given \(\mathcal F_{t-1}\). Locally around \(\theta^*\), \(m_t\) has conditional second moment of order \(O(1)\), $b_t^{\rm alg}$ and $q_t$ satisfy
\begin{equation}
\label{eq:df_component_orders}
\|b_t^{\rm alg}\|
=O(\nu_t^{\kappa_{\rm alg}})
=O(t^{-\kappa_{\rm alg}\beta}),
\qquad
\bigl\{\EE[\|q_t\|^2\mid\mathcal F_{t-1}]\bigr\}^{1/2}
=O(\sigma_t/\nu_t)
=O(t^{\beta-\zeta/2}).
\end{equation}

\subsection{Last-iterate convergence}
\label{subsec:last_iterate}
\begin{theorem}[Last-iterate convergence]
\label{thm:conv_one_two_sided}
Suppose Assumptions~\ref{as:iid_sampling}--\ref{ass:algo_config} hold and
\begin{equation}
\label{eq:as_cond_unified}
\alpha+2\kappa_{\rm alg}\beta>1,
\qquad
2\alpha+\zeta-2\beta>1.
\end{equation}
Then \(\theta_t\to\theta^*\) almost surely. If
\(\alpha\in(1/2,1)\), or if \(\alpha=1\) and \(c_\eta L>1\), then
there exist constants \(C<\infty\) and \(t_0\) such that, for all
\(t\ge t_0\),
\begin{equation}
\label{eq:mse_main_unified}
\EE\|\theta_t-\theta^*\|^2
\le
C\bigl\{
t^{-\alpha} + t^{-2\kappa_{\rm alg}\beta}
+t^{-(\alpha+\zeta-2\beta)}
\bigr\}.
\end{equation}
\end{theorem}

The conditions in \eqref{eq:as_cond_unified} control the accumulated
finite-difference bias and amplified query noise, respectively.
The three terms in \eqref{eq:mse_main_unified} represent the
stochastic-approximation error, the squared finite-difference bias,
and the amplified error from query noise.
The standard stochastic-approximation rate \(O(t^{-\alpha})\) is recovered
when \(\zeta\ge2\beta\) and \(2\kappa_{\rm alg}\beta\ge\alpha\).

For persistent query noise with \(\zeta=0\), shrinking \(\nu_t\) reduces
finite-difference bias but amplifies query noise. Balancing the last two
terms in \eqref{eq:mse_main_unified} gives \(\beta=\alpha/4\) for DF-Fwd
and \(\beta=\alpha/6\) for DF-Ind and DF-Sym, with rates
\(O(t^{-\alpha/2})\) and \(O(t^{-2\alpha/3})\), respectively.
These choices satisfy \eqref{eq:as_cond_unified} when \(\alpha>2/3\) for
DF-Fwd and \(\alpha>3/5\) for DF-Ind and DF-Sym. Thus, second-order
finite-difference bias permits a slower decay of \(\nu_t\) and less
amplification of query noise. Moreover, query reuse does not affect the
last-iterate rate order.

\subsection{Averaged inference across query-noise regimes}
\label{sec:asymp_dist_poly}

The following local moment conditions ensure covariance convergence and the
martingale Lindeberg condition.
\begin{assumption}[Inference regularity]
\label{as:clt_common_poly}
\hfill
\begin{enumerate}
\item
There exist \(r_0>0\) and \(\kappa>0\) such that\ 
\(\EE\bigl[(\mathcal E_1(s_t)\mathcal E_2(s_t))^{2+\kappa}\bigr] < \infty, \)\newline
\(
\sup_{\|\theta-\theta^*\|\le r_0}
\EE\bigl[
|\delta(\theta;s_t,r_t,s_t')|^{2+\kappa}\,(1+\mathcal E_1(s_t)^{2+\kappa})
\bigr]
<\infty,
\) and 
\(\EE[\|u_t\|^{4(2+\kappa)}] < \infty.\)
\item
The standardized query noises satisfy
\(\EE[|\epsilon_{j,t}/\sigma_t|^{4+2\kappa}]<\infty\).
\end{enumerate}
\end{assumption}
The first part controls the intrinsic TD and random-direction terms locally
around \(\theta^*\), and the second controls products of query errors.

By \eqref{eq:df_component_orders}, the sign of \(\zeta-2\beta\) determines the leading stochastic term, while a centered limit requires the
finite-difference bias to be negligible at the corresponding rate.
Table~\ref{tab:query_noise_phase_diagram} summarizes the resulting regimes.
\begin{table}[tbp]
\centering
\caption{Query-noise regimes for Polyak--Ruppert averages. }
\label{tab:query_noise_phase_diagram}
\small
\setlength{\tabcolsep}{3pt} 
\renewcommand{\arraystretch}{1.08}
\begin{tabular}{
>{\centering\arraybackslash}p{0.15\linewidth}
>{\centering\arraybackslash}p{0.12\linewidth}
>{\centering\arraybackslash}p{0.16\linewidth}
>{\centering\arraybackslash}p{0.27\linewidth}
>{\centering\arraybackslash}p{0.14\linewidth}}
\toprule
Noise regime & Lead term & Rate \(a_T\) & Additional rate conditions
& \(Q_{\rm reg}^{\rm alg}\) \\
\midrule
Fast-decaying: \(\zeta>2\beta\)
& \(m_t\) & \(\sqrt T\) & \(\kappa_{\rm alg}\beta>1/2\) & \(Q\) \\
Boundary: \(\zeta=2\beta>0\)
& \(m_t + q_t\) & \(\sqrt T\) & \(\kappa_{\rm alg}\beta>1/2\)
& \(Q_{\rm bdry}^{\rm alg}\) \\
Intermediate: \(0<\zeta<2\beta\)
& \(q_t\) & \(T^{(1-2\beta+\zeta)/2}\)
& \(2\alpha+\zeta-2\beta>1\), and
  \(\kappa_{\rm alg}\beta>(1-2\beta+\zeta)/2\)
& \(Q_{\rm int}^{\rm alg}\) \\
Persistent: \(\zeta=0\)
& \(q_t\) & \(T^{(1-2\beta)/2}\)
& \(2\alpha-2\beta>1\), and
  \(\kappa_{\rm alg}\beta>(1-2\beta)/2\)
& \(Q_{\rm pers}^{\rm alg}\) \\
\bottomrule
\end{tabular}
\end{table}
In the fast-decaying regime,
\begin{equation}
\label{eq:Q}
Q
=
\EE\Bigl[(\delta_t^*)^2
\bigl(\nabla_\theta V(s_t;\theta^*)^\top u_t\bigr)^2u_tu_t^\top\Bigr].
\end{equation}
The algorithm-specific covariances in the other regimes are
\begin{equation}
\label{eq:Q_bdry}
\begin{aligned}
Q_{\rm bdry}^{\rm Ind}=Q_{\rm bdry}^{\rm Sym}
=Q+\frac{c_\sigma\EE[(\delta_t^*)^2]}{2c_\nu^2}I_d,
\qquad
&Q_{\rm bdry}^{\rm Fwd}
=Q+\frac{2c_\sigma\EE[(\delta_t^*)^2]}{c_\nu^2}I_d,
\\
Q_{\rm int}^{\rm Ind}=Q_{\rm int}^{\rm Sym}
=\frac{c_\sigma\EE[(\delta_t^*)^2]}
{2c_\nu^2(1+2\beta-\zeta)}I_d,
\qquad
&Q_{\rm int}^{\rm Fwd}
=\frac{2c_\sigma\EE[(\delta_t^*)^2]}
{c_\nu^2(1+2\beta-\zeta)}I_d.
\end{aligned}
\end{equation}
For the persistent regime, write
\(\mu_4:=\EE[(\epsilon_{j,t}/\sigma_t)^4]\). Then
\begin{equation}
\label{eq:Q_pers}
\begin{aligned}
Q_{\rm pers}^{\rm Ind}
&=\frac{\frac{c_\sigma}{2}\EE[(\delta_t^*)^2]
+\frac{c_\sigma^2}{2}(1+\gamma^2)}
{c_\nu^2(1+2\beta)}I_d,
\quad Q_{\rm pers}^{\rm Fwd}
=\frac{2c_\sigma\EE[(\delta_t^*)^2]
+c_\sigma^2(2\gamma^2+1+\mu_4)}
{c_\nu^2(1+2\beta)}I_d,
\\
Q_{\rm pers}^{\rm Sym}
&=\frac{\frac{c_\sigma}{2}\EE[(\delta_t^*)^2]
+c_\sigma^2\{\frac{\gamma^2}{2}
+\frac{\mu_4-1}{8}\}}
{c_\nu^2(1+2\beta)}I_d.
\end{aligned}
\end{equation}
The terms proportional to \(c_\sigma\) arise from the TD residual multiplying the noise of the gradient queries. The terms proportional to
\(c_\sigma^2\) arise from multiplying the query noises in the residual and
the gradient surrogate. Their coefficients depend on query reuse and the
standardized fourth moment of the query noise.

\begin{theorem}[Unified CLT across query-noise regimes]
\label{thm:clt_unified}
Suppose Assumptions~\ref{as:iid_sampling}--\ref{as:clt_common_poly} hold and
\(\alpha\in(1/2,1)\). For each DF-TD algorithm and each regime in
Table~\ref{tab:query_noise_phase_diagram} satisfying its rate conditions,
\[
a_T(\bar\theta_T-\theta^*)
\Longrightarrow
\mathcal N\bigl(
0,\;
(-J^*)^{-1}Q_{\rm reg}^{\rm alg}(-J^*)^{-\top}
\bigr),
\]
where \({\rm alg}\) denotes the algorithm and \({\rm reg}\) the query-noise regime.
\end{theorem}

\begin{remark}[Instruments and query design]
\label{rem:instruments}
Under misspecification, instrument choice can change the 
calibration target, as discussed in Section~\ref{sec:background}. 
Under correct specification, the value-matching parameter
is a common root for all admissible instruments, although
their statistical efficiency may differ. 
We use the TD instrument \(W=\nabla_\theta V\). With this instrument fixed,
random-direction gradient approximation and query noise
introduce two further sources of variability.
The resulting asymptotic covariances satisfy
\(
\Sigma_{\rm eff}
\preceq
(-J^*)^{-1}\Sigma_{\rm TD}(-J^*)^{-\top}
\preceq
(-J^*)^{-1}Q(-J^*)^{-\top}
\preceq
(-J^*)^{-1}Q_{\rm bdry}^{\rm alg}(-J^*)^{-\top},
\)
where \(\Sigma_{\rm eff}\) is the efficiency bound obtained
by optimally choosing the instrument \(W\)
\citep{Chamberlain1987}.
The successive comparisons reflect instrument choice,
random-direction approximation \citep{ChenLaiLiZhang2024}, and query noise in the
boundary regime, respectively.
In the fast-decaying regime, all three designs share the same limiting
score covariance \(Q\). 
When query noise is not negligible, query design does not
change the TD instrument but determines how query noise
contributes to the limiting covariance. 
 Propositions S1 and S2 in the Appendix state and prove these claims.
\end{remark}

\begin{remark}[Effect of query reuse under persistent noise]
\label{rem:reuse_persistent}
DF-Ind and DF-Sym use the same two-sided gradient surrogate. DF-Ind uses
an independent central query in the residual, whereas DF-Sym uses the
midpoint of the same perturbed pair. The contribution of the queries at the current state to
\(N_{\delta,t}N_{\nabla,t}\) is therefore
\[
-\frac{\epsilon_{2,t}(\epsilon_{3,t}-\epsilon_{4,t})}{2\nu_t}u_t
\ \text{for DF-Ind},
\qquad
-\frac{\epsilon_{3,t}^2-\epsilon_{4,t}^2}{4\nu_t}u_t
\ \text{for DF-Sym}.
\]
For \(\zeta>0\), these noise products are asymptotically lower order,
and the two designs have the same leading contribution from query noise.
At \(\zeta=0\), they contribute at first order, giving
\[
Q_{\rm pers}^{\rm Sym}-Q_{\rm pers}^{\rm Ind}
=\frac{c_\sigma^2(\mu_4-5)}{8c_\nu^2(1+2\beta)}I_d.
\]
Hence DF-Sym has a smaller asymptotic covariance than DF-Ind when \(\mu_4<5\).
\end{remark}

The limit theorem for the averaged parameter estimate also yields inference for the calibrated value
\(V(s;\theta^*)\).
\begin{corollary}[Limit for the calibrated value]
\label{cor:value_inference_mc}
Let \(s\) be the reporting state defined in
Section~\ref{sec:background}, and suppose that
\(\theta\mapsto V(s;\theta)\) is continuously differentiable at
\(\theta^*\). Under the corresponding regime of
Theorem~\ref{thm:clt_unified},
\[
a_T\bigl(V(s;\bar\theta_T)-V(s;\theta^*)\bigr)
\Longrightarrow
\mathcal N\bigl(
0,\;
\nabla_\theta V(s;\theta^*)^\top
(-J^*)^{-1}Q_{\rm reg}^{\rm alg}(-J^*)^{-\top}
\nabla_\theta V(s;\theta^*)
\bigr).
\]
\end{corollary}

\subsection{Fixed perturbation radius under correct specification}
\label{subsec:perfect_simulators}

For DF-Ind and DF-Fwd, the bias of gradient surrogate in Lemma~\ref{lem:gradientbias} enters the mean TD update weighted by the Bellman discrepancy. 
For fixed \(\theta\) and radius \(\nu>0\), let
\(G_\nu^{\rm alg}(\theta)\) denote the expectation of the exact-query
increment \(g_t^{\rm ex}\), so that
\(b_t^{\rm alg}=G_{\nu_t}^{\rm alg}(\theta_{t-1})-G(\theta_{t-1})\).

\begin{proposition}[Finite-difference bias and Bellman discrepancy]
\label{prop:bias_discrepancy}
Under Assumption~\ref{ass:regularity}(2) and the conditions and constants \(C_1,C_2\) of
Lemma~\ref{lem:gradientbias}, for every \(\theta\) and \(\nu>0\),
\[
\begin{aligned}
\bigl\|G_\nu^{\rm Fwd}(\theta)-G(\theta)\bigr\|&
\le C_1\,\nu\,\|\varepsilon_{\rm B}(\cdot;\theta)\|_\mu\|\mathcal E_2\|_\mu,\quad
\bigl\|G_\nu^{\rm Ind}(\theta)-G(\theta)\bigr\| 
\le C_2\,\nu^2\,\|\varepsilon_{\rm B}(\cdot;\theta)\|_\mu\|\mathcal E_3\|_\mu,\\
\bigl\|G_\nu^{\rm Sym}(\theta)-G(\theta)\bigr\|
&\le C_2\,\nu^2\,\|\varepsilon_{\rm B}(\cdot;\theta)\|_\mu\|\mathcal E_3\|_\mu+\frac{\nu^2}{2}\EE[\mathcal E_1(s)\mathcal E_2(s)]\,\EE\|u_t\|^4.
\end{aligned}
\]
Moreover
\(\|\varepsilon_{\rm B}(\cdot;\theta)\|_\mu
\le\|\varepsilon_{\rm B}(\cdot;\theta^*)\|_\mu
+(1+\gamma)\|\mathcal E_1\|_\mu\|\theta-\theta^*\|\).
\end{proposition}
Combining the preceding bounds gives
\(\|b_t^{\rm alg}\|\le C\nu_t^{\kappa_{\rm alg}}(1+\|\theta_{t-1}-\theta^*\|)\),
as used in Theorems~\ref{thm:conv_one_two_sided} and
\ref{thm:clt_unified}.
These bounds also control the displacement of the root at a fixed radius. Define
\(K_{\rm alg}:=C_{\rm alg}(1+\gamma)
\|\mathcal E_1\|_\mu\|\mathcal E_{\kappa_{\rm alg}+1}\|_\mu\),
with \(C_{\rm Fwd}=C_1\) and
\(C_{\rm Ind}=C_{\rm Sym}=C_2\), and
\(\nu_{\rm alg}:=(L/K_{\rm alg})^{1/\kappa_{\rm alg}}\).

\begin{corollary}[Root of the mean field at a fixed radius]
\label{cor:fixed_radius_root}
Under Assumption~\ref{ass:regularity}, let \(\nu<\nu_{\rm alg}\) and suppose
that \(G_\nu^{\rm alg}\) has a root \(\theta_\nu^{\rm alg}\). Then
\[
\|\theta_\nu^{\rm alg}-\theta^*\|
\le
\frac{C_{\rm alg}\,\nu^{\kappa_{\rm alg}}\|\mathcal E_{\kappa_{\rm alg}+1}\|_\mu\,\|\varepsilon_{\rm B}(\cdot;\theta^*)\|_\mu
+\mathbf 1\{{\rm alg}={\rm Sym}\}\,\frac{\nu^2}{2}\EE[\mathcal E_1(s)\mathcal E_2(s)]\,\EE\|u_t\|^4}
{L-K_{\rm alg}\nu^{\kappa_{\rm alg}}}.
\]
\end{corollary}

Under correct specification the first term of this bound vanishes, so
\(\theta^*\) remains a root of \(G_\nu^{\rm alg}\) for DF-Ind and DF-Fwd at
any fixed radius, regardless of finite-difference accuracy, whereas DF-Sym
may shift the root through its midpoint residual.

For DF-Ind and DF-Fwd,
Lemma~S1 in the Appendix
shows that \(G_\nu^{\rm alg}\) satisfies
Assumption~\ref{ass:regularity}(1) with constant
\(L_\nu:=L-K_{\rm alg}\nu^{\kappa_{\rm alg}}\)
whenever \(0<\nu<\nu_{\rm alg}\).
Define
\(J_\nu^{\rm alg}:=\nabla G_\nu^{\rm alg}(\theta^*)\) and
\(Q_\nu^{\rm alg}:=
\EE[(\delta_t^*)^2
\hat\nabla_t^{\rm ex}\hat\nabla_t^{{\rm ex}\top}]\),
where the exact-query surrogate is evaluated at
\(\theta^*\) with radius \(\nu\).

\begin{theorem}[Calibration at a fixed radius under correct specification]
\label{thm:fixed_radius}
Suppose the simulator is correctly specified with value-matching parameter \(\theta^*\), that
Assumptions~\ref{as:iid_sampling}--\ref{as:clt_common_poly} hold with the
radius schedule in \eqref{eq:polyrate} replaced by a constant
\(\nu_t\equiv\nu\in(0,\nu_{\rm alg})\), and that
\({\rm alg}\in\{{\rm Ind},{\rm Fwd}\}\).
\begin{enumerate}
\item[(1)] If \(\alpha\in(1/2,1)\), or if \(\alpha=1\) and
\(c_\eta L_\nu>1\), then for any \(\zeta\ge0\), \(\theta_t\to\theta^*\)
almost surely and \(\EE\|\theta_t-\theta^*\|^2=O(t^{-\alpha})\).
\item[(2)] If \(\alpha\in(1/2,1)\) and \(\zeta>0\), then
\(\sqrt T(\bar\theta_T-\theta^*)\Rightarrow
\mathcal N(0,(-J_\nu^{\rm alg})^{-1}Q_\nu^{\rm alg}(-J_\nu^{\rm alg})^{-\top})\).
\end{enumerate}
\end{theorem}
Under correct specification, the bias conditions
\(\alpha+2\kappa_{\rm alg}\beta>1\) and
\(\kappa_{\rm alg}\beta>1/2\) in Theorems~\ref{thm:conv_one_two_sided}
and \ref{thm:clt_unified} are no longer needed.
With a fixed radius, the amplification of query noise no longer grows over time, and the fast-decaying condition $\zeta>2\beta$ reduces to $\zeta > 0$.
When finite differencing introduces no approximation error in the
TD increment, the results of
Theorem~\ref{thm:fixed_radius} also hold for all three designs
with any fixed \(\nu>0\), even under misspecification.
In this case, \(G_\nu^{\rm alg}=G\),
\(J_\nu^{\rm alg}=J^*\), and \(Q_\nu^{\rm alg}=Q\).

\subsection{Design trade-offs under budgets of transitions and rollouts}
\label{sec:budget}

The preceding results compare the query designs in terms of query cost per update, the finite-difference
accuracy, and the covariance contribution of query noise.
Table~\ref{tab:algorithm_comparison} collects these trade-offs, together
with two properties of the target discussed below.
We refer to \(\kappa_{\rm alg}=1\) and \(\kappa_{\rm alg}=2\) as the
first- and second-order cases.

\begin{table}[tbp]
\centering
\caption{Query cost, finite-difference accuracy, leading effects of query noise, and target preservation across the three DF-TD designs. The gradient-noise factors multiply \((\sigma_t^2/\nu_t^2)I_d\) in \(\operatorname{Cov}(N_{\nabla,t}\!\mid\!\mathcal F_{t-1})\). The product-noise factors multiply \(c_\sigma^2\{c_\nu^2(1+2\beta)\}^{-1}I_d\) in \(Q_{\rm pers}^{\rm alg}\). The root-shift bounds follow from Corollary~\ref{cor:fixed_radius_root}. The last row assumes query variance \(v(s,\theta)/M\) at fixed replication \(M\) and follows 
from Proposition S3 in the Appendix.
}
\label{tab:algorithm_comparison}
\small
\setlength{\tabcolsep}{6pt}
\renewcommand{\arraystretch}{1.25}
\begin{tabular*}{\linewidth}{@{\extracolsep{\fill}}lccc@{}}
\toprule
Quantity & DF-Ind & DF-Fwd & DF-Sym \\
\midrule
Fresh queries per update \(L_{\rm alg}\)
& \(4\) & \(3\) & \(3\) \\

Bias order \(\kappa_{\rm alg}\)
& \(2\) & \(1\) & \(2\) \\

Gradient-noise factor
& \(\dfrac12\) & \(2\) & \(\dfrac12\) \\

Covariance in the fast-decaying regime
& \(Q\) & \(Q\) & \(Q\) \\

Product-noise factor, persistent regime
& \(\dfrac{1+\gamma^2}{2}\)
& \(2\gamma^2+1+\mu_4\)
& \(\dfrac{\gamma^2}{2}+\dfrac{\mu_4-1}{8}\) \\

Root shift under correct specification at fixed \(\nu\)
& none & none & \(O(\nu^2)\) \\

Bias from heteroscedastic query noise as \(\nu\downarrow0\)
& none & none & \(-\dfrac{\mathbb E[\nabla_\theta v(s,\theta)]}{2M}\) \\
\bottomrule
\end{tabular*}
\end{table}

In the fast-decaying regime, all three designs share the score covariance
\(Q\). DF-Sym matches DF-Ind in two-sided accuracy and in the noise level of the gradient queries, with one fewer query. DF-Fwd also uses
three queries, but its one-sided surrogate has four times the gradient-noise factor. Under persistent noise, the covariance
comparison in Remark~\ref{rem:reuse_persistent} provides an additional
criterion for choosing between DF-Ind and DF-Sym.

\begin{figure}[tbp]
\centering
\includegraphics[width=0.82\linewidth]
{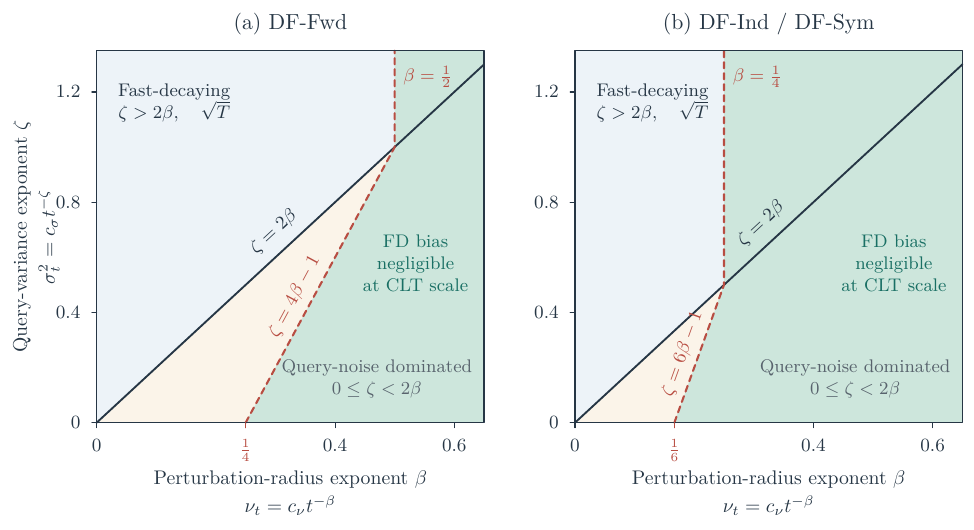}
\caption{Tuning geometry for averaged DF-TD inference when the
finite-difference bias at the root does not vanish. The solid line \(\zeta=2\beta\) separates the fast-decaying region from the region dominated by query noise. Dashed lines give the boundaries for the finite-difference bias in the first- and second-order cases. Table~\ref{tab:query_noise_phase_diagram} gives the
exact rates and the additional \(\alpha\)-dependent conditions.}
\label{fig:phase_diagram}
\end{figure}

The finite-difference order also determines how the perturbation radius and
query precision can be scheduled for centered inference.
Figure~\ref{fig:phase_diagram} gives a geometric representation of the rate
conditions in Table~\ref{tab:query_noise_phase_diagram} when the
finite-difference bias at the root does not vanish.
Compared with the first-order case, second-order finite-difference accuracy
admits a wider set of schedules for the perturbation radius and the query precision for centered
inference. In the fast-decaying regime, the threshold for the finite-difference bias decreases
from \(\beta>1/2\) to \(\beta>1/4\). At \(\zeta=0\), the corresponding
thresholds are \(1/4\) and \(1/6\). Second-order accuracy therefore permits a
more slowly shrinking perturbation radius and slower growth of the
replication count \(M_t\).

The radius and replication choices can also affect the calibration target, as summarized in the last two rows of Table~\ref{tab:algorithm_comparison}. Under correct specification, DF-Ind and DF-Fwd preserve \(\theta^*\) at a sufficiently small fixed radius by Theorem~\ref{thm:fixed_radius}. DF-Sym can shift the root through its midpoint residual unless \(V(s;\theta)\) is affine in \(\theta\). With query variance \(v(s,\theta)/M\) at fixed replication \(M\), query reuse can also bias the DF-Sym mean field unless  \(v(s,\theta)\) is independent of \(\theta\). 
See Proposition S3 for the small-radius limit of this bias and a bound on the resulting root shift.

The requirements on the perturbation radius and query precision determine the simulator rollout cost for a given number of target-system transitions. Each TD update uses one transition of the target system. When each returned value averages \(M_t\asymp t^\zeta\) independent simulator rollouts, \eqref{eq:main_rollout_budget} gives the cost of \(T\) updates as
\(
C_T^{\rm main}
\asymp L_{\rm alg}T^{1+\zeta}.
\)

For inference, the question is how many simulator rollouts are needed to make query noise negligible.
Under the fast-decaying conditions in
Table~\ref{tab:query_noise_phase_diagram}, \(\zeta>2\beta\) and
\(\kappa_{\rm alg}\beta>1/2\) require \(\zeta>1\) in the first-order case
and \(\zeta>1/2\) in the second-order case. Let \(\epsilon>0\) denote
an arbitrarily small, fixed margin above these thresholds for
\(\zeta\). Choosing \(\zeta=1+\epsilon\) or \(\zeta=1/2+\epsilon\),
respectively, supports centered root-\(T\) inference with rollout budgets of order \(T^{2+\epsilon}\) or \(T^{3/2+\epsilon}\).

When performance is indexed only by the rollout budget, the root-\(T\) rate becomes
\(T^{-1/2}\asymp(C_T^{\rm main})^{-1/\{2(1+\zeta)\}}\).
As \(\zeta\) approaches \(1\) and \(1/2\) from
above in the first- and second-order cases, respectively,
the exponents approach \(1/4\) and \(1/3\), matching the Kiefer-Wolfowitz rates
with independent evaluation noise \citep{Fabian1967,DipponRenz1997}. The
persistent regime of Table~\ref{tab:query_noise_phase_diagram} reaches the
same exponents with fixed rollout replication, so replication changes the
rate in \(T\) but not the frontier in the rollout budget.
By contrast, under correct specification, both DF-Ind and DF-Fwd
can use a fixed radius and retain centered root-\(T\) inference
with \(\zeta>0\), as established in
Theorem~\ref{thm:fixed_radius}.
The exponent \(1/\{2(1+\zeta)\}\) then approaches \(1/2\)
as \(\zeta\downarrow0\). The preferred design therefore
depends on how accurate the  simulator is believed to be.

\section{Few-replicate Online Bootstrap}
\label{sec:online_bootstrap}

\begingroup
\setstretch{1.0}
\begin{algorithm}[tbp]
\caption{Few-replicate online bootstrap for DF-TD}
\label{alg:bootstrap}
\small
\SetAlgoLined
\DontPrintSemicolon
\KwIn{\(\theta_0\), \(\{\eta_t\}\), \(\{\nu_t\}\), horizon \(T\),
bootstrap size \(B\ge2\), level \(1-\tau\), and a DF-TD variant}
\KwOut{Main and bootstrap averages and coordinatewise intervals}
Initialize \(\bar\theta_0\gets\theta_0\); for each \(b\in[B]\), set
\(\theta_0^{(b)}\gets\theta_0\) and
\(\bar\theta_0^{(b)}\gets\theta_0\)\;
\For{\(t=1,\ldots,T\)}{
    Sample \((s_t,r_t,s_t')\) and \(u_t\sim\mathcal P_u\)\;
    Run one DF-TD step of the main path using Algorithm~\ref{alg:zo_td}\;
    \For{\(b=1,\ldots,B\)}{
        Sample \(W_t^{(b)}\) according to
        Assumption~\ref{as:boot_weight}\;
        Reuse \((s_t,r_t,s_t')\) and \(u_t\), draw fresh query
        noises, and compute \(\hat g_t^{(b)}\) at
        \(\theta_{t-1}^{(b)}\)\;
        \(\theta_t^{(b)}
        \gets\theta_{t-1}^{(b)}
        +\eta_tW_t^{(b)}\hat g_t^{(b)}\)\;
        \(\bar\theta_t^{(b)}
        \gets\frac{t-1}{t}\bar\theta_{t-1}^{(b)}
        +\frac1t\theta_t^{(b)}\)\;
    }
    \(\bar\theta_{t,B}^{\rm boot}
    \gets B^{-1}\sum_{b=1}^B\bar\theta_t^{(b)}\)\;
    \For{\(i = 1, \ldots,d\)}{
        \(S_{t,i}^2
        \gets\frac{t}{B-1}\sum_{b=1}^B
        \big\{(\bar\theta_t^{(b)})_i
        -(\bar\theta_{t,B}^{\rm boot})_i\big\}^2\)\;
        \(\CI_{t,i}^{\rm boot}
        \gets
        [(\bar\theta_t)_i
        \pm t_{B-1,1-\tau/2}S_{t,i}/\sqrt t]\)\;
    }
}
\end{algorithm}
\endgroup
Because each bootstrap path must be evaluated at its own parameter, every
path costs as many simulator rollouts as the main path. Thus, \(B\)
directly controls simulation cost, motivating inference with a fixed, small
number of bootstrap paths.

We develop an online multiplier bootstrap for DF-TD using a fixed number
\(B\ge2\) of bootstrap paths in the fast-decaying regime.
Each path shares the observed transition and perturbation direction with
the main path, but uses an independent multiplier and fresh local
value queries at its own parameter, which introduce an
additional source of randomness.

All paths are updated in one pass without storing the sample trajectory. As in \eqref{eq:main_rollout_budget}, the total training budget is \((B+1)L_{\rm alg}\sum_{t=1}^T M_t=(B+1)C_T^{\rm main}\). \begin{assumption}[Bootstrap weights and query streams] \label{as:boot_weight} For fixed $B$, the array $\{W_t^{(b)}:t\ge1,b\in\{1, \dots, B\}\}$ is independent across $(t,b)$, with $\EE[W_t^{(b)}]=1$, $\Var(W_t^{(b)})=1$, and, for some $\kappa_W>0$, $\sup_{t,b}\EE\bigl[|W_t^{(b)}|^{2+\kappa_W}\bigr]<\infty$. The weight array is independent of $\sigma(\cup_T\mathcal F_T)$ and of all bootstrap query noises. For each $b$, the bootstrap query noises are an independent copy of the original sequence of query noises and are independent across queries, time, and replicates. \end{assumption} Examples of weights satisfying Assumption~\ref{as:boot_weight} include $\mathrm{Exp}(1)$ and $\mathrm{Unif}(1-\sqrt{3},1+\sqrt{3})$.

Algorithm~\ref{alg:bootstrap} gives the resulting one-pass procedure.
The bootstrap dispersion is computed around the bootstrap sample mean,
whereas the reported interval is centered at the main estimator. 
Theorem~\ref{thm:bootstrap_joint_limit} gives the conditional and joint
limits that fixed-\(B\) inference requires.
\begin{theorem}[Few-replicate bootstrap joint limit]
\label{thm:bootstrap_joint_limit}
Under the assumptions for the fast-decaying regime in Theorem~\ref{thm:clt_unified} and Assumption~\ref{as:boot_weight}, let
\(
\Sigma_{\rm std}:=(-J^*)^{-1}Q(-J^*)^{-\top}.
\)

\textnormal{(a) Conditional validity.}
For each fixed \(b\in\{1,\ldots,B\}\), conditionally on \(\mathcal F_T\),
\[
\sqrt{T}\big(\bar\theta_T^{(b)}-\bar\theta_T\big)
\Longrightarrow
\mathcal N(0,\Sigma_{\rm std}).
\]

\textnormal{(b) Joint fixed-\(B\) limit.}
For any fixed \(B\ge2\),
\[
\sqrt{T}\big(\bar\theta_T-\theta^*,
\bar\theta_T^{(1)}-\bar\theta_T,\ldots,
\bar\theta_T^{(B)}-\bar\theta_T\big)
\Longrightarrow
(Z_0,Z_1,\ldots,Z_B),
\]
where \(Z_0,Z_1,\ldots,Z_B\) are independent and identically distributed as
\(\mathcal N(0,\Sigma_{\rm std})\).
\end{theorem}
The joint limit in Theorem~\ref{thm:bootstrap_joint_limit}(b) allows the
bootstrap dispersion to Studentize the main estimation error. Independence
of the Gaussian components and their common covariance yield the
\(t_{B-1}\) pivot in Corollary~\ref{cor:cheap_bootstrap_t}. For a coordinate
\(i\), let \(S_{T,i}\) be the dispersion computed by
Algorithm~\ref{alg:bootstrap} at \(t=T\), and set the ratio
\(\sqrt T\{(\bar\theta_T)_i-(\theta^*)_i\}/S_{T,i}\) to zero on the event
\(S_{T,i}=0\).
\begin{corollary}
\label{cor:cheap_bootstrap_t}
Suppose the assumptions of Theorem~\ref{thm:bootstrap_joint_limit}(b) hold,
and let \(i\in \{1,\ldots,d\}\) be fixed with \((\Sigma_{\rm std})_{ii}>0\). Then
\[
\frac{\sqrt{T}\big((\bar\theta_T)_i-(\theta^*)_i\big)}{S_{T,i}}
\Longrightarrow t_{B-1}.
\]
Consequently, for any $\tau\in(0,1)$,
\[
\PP\Bigl(
(\theta^*)_i
\in
\Bigl[
(\bar\theta_T)_i
-
\frac{t_{B-1,1-\tau/2}}{\sqrt{T}}\,S_{T,i},\,
(\bar\theta_T)_i
+
\frac{t_{B-1,1-\tau/2}}{\sqrt{T}}\,S_{T,i}
\Bigr]
\Bigr)
\to 1-\tau.
\]
\end{corollary}
The interval can be reported at any prespecified deterministic checkpoint. Under the fixed-radius
conditions of Theorem~\ref{thm:fixed_radius}(2),
Theorem~\ref{thm:bootstrap_joint_limit} and
Corollary~\ref{cor:cheap_bootstrap_t} remain valid, with \(\Sigma_{\rm std}\)
replaced by \((-J_\nu^{\rm alg})^{-1}Q_\nu^{\rm alg}(-J_\nu^{\rm alg})^{-\top}\).
\begin{remark}[Inference for the value at a fixed state]
\label{rem:bootstrap_value_t}
The same construction can be used for value reporting at \(s\). Apply the joint delta method to
\(\theta\mapsto V(s;\theta)\), and define
\[
S_{T,s}^2
=
\frac{T}{B-1}\sum_{b=1}^B
\Bigl\{V(s;\bar\theta_T^{(b)})
-\frac1B\sum_{b'=1}^BV(s;\bar\theta_T^{(b')})\Bigr\}^2.
\]
If \(\theta\mapsto V(s;\theta)\) is continuously differentiable at \(\theta^*\) and
\(\nabla_\theta V(s;\theta^*)^\top\Sigma_{\rm std}
\nabla_\theta V(s;\theta^*)>0\), then
\[
\frac{\sqrt T\{V(s;\bar\theta_T)-V(s;\theta^*)\}}{S_{T,s}}
\Longrightarrow t_{B-1}.
\]
Hence a level-\((1-\tau)\) interval for \(V(s;\theta^*)\) is 
\(\bigl[
V(s;\bar\theta_T)
-
\frac{t_{B-1,1-\tau/2}}{\sqrt T}\,S_{T,s},
V(s;\bar\theta_T)
+
\frac{t_{B-1,1-\tau/2}}{\sqrt T}\,S_{T,s}\bigr].
\)

If the values used to construct this interval are reported from fresh
simulator rollouts, the interval retains its asymptotic coverage when the additional Monte
Carlo error and truncation bias are \(o_p(T^{-1/2})\). 
Appendix discusses these terminal reporting errors.
\end{remark}

The results above assume independent one-step samples. Along a
trajectory, the score at the target,
\(\psi_t^*:=\delta_t^*\{\nabla_\theta V(s_t;\theta^*)^\top u_t\}u_t\), is in general dependent
over time. Its lag-zero covariance is \(Q\), because \(\psi_t^*\) is centered and each
transition along the stationary trajectory has the law in
Assumption~\ref{as:iid_sampling}. Write
\(\Omega_{\rm M}=Q+\sum_{\ell\ne0}
\operatorname{Cov}(\psi_0^*,\psi_\ell^*)\) for its long-run covariance, and let
\(\Sigma_{\rm M}:=(-J^*)^{-1}\Omega_{\rm M}(-J^*)^{-\top}\). 
Theorem S3 in the Appendix gives the full conditions and the proof of the following statement.

\begin{proposition}[Bootstrap under trajectory sampling]
\label{prop:markov_bootstrap_main}
Suppose the observations are generated along a stationary, uniformly geometrically ergodic
Markov Chain. Assume the conditions of the fast-decaying regime and the Markov regularity conditions
of Appendix hold, and
that all \(B+1\) paths of Algorithm~\ref{alg:bootstrap} remain in a
fixed compact set almost surely. Then
\(\sqrt T(\bar\theta_T-\theta^*)\Rightarrow\mathcal N(0,\Sigma_{\rm M})\),
while the
bootstrap deviations \(\sqrt T(\bar\theta_T^{(b)}-\bar\theta_T)\Rightarrow\mathcal N(0,\Sigma_{\rm std})\) jointly across \(b\le B\) conditionally on the data.
For any coordinate \(i\), the \(t_{B-1}\) intervals in Corollary~\ref{cor:cheap_bootstrap_t} are calibrated if and only if  \((\Sigma_{\rm M})_{ii}=(\Sigma_{\rm std})_{ii}\). A
sufficient condition is that \(\psi_t^*\)
be a martingale difference,
which for TD calibration is equivalent to
\(\varepsilon_{\rm B}(s_t;\theta^*)\nabla_\theta V(s_t;\theta^*)=0\)
almost surely.
\end{proposition}

The Poisson equation for the chain explains the two limits. It splits \(\psi_t^*\) into a martingale difference and a telescoping remainder \(\rho_{t-1}-\rho_t\). Along the main path, the remainder is weighted by the slowly varying Polyak--Ruppert weights, making its sum \(o(\sqrt T)\). Along a bootstrap path the remainder is weighted by \(W_t^{(b)}-1\), which is not slowly varying. After summation by parts it becomes \(\sum_t(W_{t+1}^{(b)}-W_t^{(b)})\rho_t\), whose conditional variance can be of order \(T\). Conditionally on the data, \(\Var\{\frac{1}{\sqrt T}\sum_t(W_t^{(b)}-1)\psi_t^*\}=\frac{1}{T }\sum_t\psi_t^*\psi_t^{*\top}\to Q\), and the bootstrap limit therefore has the lag-zero covariance.
In contrast to Theorem~4.2 of \citet{RamprasadEtAl2023}, which concludes that the online multiplier bootstrap for linear TD is distributionally consistent along uniformly ergodic Markov trajectories without requiring the score to be a martingale difference, we show that this bootstrap can be inconsistent when this score has temporal dependence.

The martingale condition in Proposition~\ref{prop:markov_bootstrap_main} is equivalent to the vanishing of the discrepancy term in Remark~\ref{rem:td_variance_decomposition}. A correctly specified simulator with a unique TD root satisfies the condition. If the condition fails, the intervals need not be calibrated, as the following example shows. Consider a two-state chain \(s_t\in\{-1,1\}\) with uniform stationary law and \(\EE[s_{t+1}\mid s_t]=\varrho s_t\) for some \(\varrho\in(0,1)\), reward \(r_t=s_t\), and the one-parameter linear family \(V(s;\theta)=\theta\). Take \(u_t\) to be Rademacher. This is linear TD with a constant feature. Here \(\theta^*=0\), the score is \(\psi_t^*=s_t\), the discrepancy \(\varepsilon_{\rm B}(s;0)=s\) is autocorrelated along the chain, and \(Q=1\) while \(\Omega_{\rm M}=(1+\varrho)/(1-\varrho)\). The bootstrap standard error is therefore multiplied by \(\{(1-\varrho)/(1+\varrho)\}^{1/2}\), and a nominal \(1-\tau\) interval has limiting coverage \(\PP(|t_{B-1}|\le\{(1-\varrho)/(1+\varrho)\}^{1/2}t_{B-1,1-\tau/2})\). Random scaling (e.g., \citealp{lee2022randomscaling}) may avoid this mismatch by preserving the Poisson cancellation along the main path at the cost of nonstandard critical values and wider intervals than methods based on covariance estimates. We leave its extension to DF-TD for future work.

\section{Correcting the Calibrated Value}
\label{sec:discrepancy_correction}

For a misspecified simulator the calibrated value differs from the policy
value by
\(\Delta(s):=V^\pi(s)-V(s;\theta^*)\). Given \eqref{eq:value_discrepancy_resolvent}, the value discrepancy $\Delta(s)$ solves a
Bellman equation,
\begin{equation}
\label{eq:discrepancy_bellman}
\Delta(s)
=
\varepsilon_{\rm B}(s;\theta^*)
+\gamma(P^\pi\Delta)(s)
=
\EE\bigl[
\delta(\theta^*;s,r,s')+\gamma\Delta(s')
\mid s
\bigr].
\end{equation}
This suggests augmenting the DF-TD calibration in
Algorithm~\ref{alg:zo_td} with a second-stage
TD recursion that uses the TD residuals already computed to estimate the \(\Delta\).

To approximate \(\Delta\), let \(\chi:\mathcal S\to\mathbb R^p\) be a fixed feature map, chosen to capture discrepancies left by simulator calibration. Define
\(A_\chi:=\EE[\chi(s)\{\chi(s)-\gamma\chi(s')\}^\top]\),
\(b_\chi(\theta):=\EE[\delta(\theta;s,r,s')\chi(s)]\), and, when
\(A_\chi\) is nonsingular, \(\omega^*:=A_\chi^{-1}b_\chi(\theta^*)\). The
corrected value is \(V^{\rm c}(s):=V(s;\theta^*)+\chi(s)^\top\omega^*\). In particular,
using only the gradient features
\(\nabla_\theta V(\cdot;\theta^*)\) would give
\(b_\chi(\theta^*)=G(\theta^*)=0\), and hence no correction.
The following proposition relates the remaining error to how
well the span of \(\chi\) approximates \(\Delta\).

\begin{proposition}[Projected value discrepancy]
\label{prop:discrepancy_projection}
If \(\Sigma_\chi:=\EE[\chi(s)\chi(s)^\top]\succ0\) and
\(\EE\|\chi(s)\|^2<\infty\), then \((A_\chi+A_\chi^{\top})/2\succeq(1-\gamma)\Sigma_\chi\),
\(\chi^\top\omega^*\) is the unique fixed point of
\(f\mapsto\Pi_\chi\{\varepsilon_{\rm B}(\cdot;\theta^*)+\gamma P^\pi f\}\)
with \(\Pi_\chi\) the \(L_2(\mu)\)-projection onto the span of \(\chi\),
and
\(\|\chi^\top\omega^*-\Delta\|_\mu
\le(1-\gamma)^{-1}\|\Pi_\chi\Delta-\Delta\|_\mu\).
If \(\Delta=\chi^\top\omega_0\) \(\mu\)-almost everywhere, then
\(\omega^*=\omega_0\) and \(V^{\rm c}=V^\pi\) \(\mu\)-almost everywhere.
\end{proposition}

The second stage estimates \(\omega^*\) through the following
recursion, which runs alongside the first stage without
additional simulator queries:
\begin{equation}
\label{eq:second_stage}
\omega_t
=\omega_{t-1}
+\eta_t\,
\bigl\{\hat\delta_t+\gamma\chi(s_t')^\top\omega_{t-1}
-\chi(s_t)^\top\omega_{t-1}\bigr\}\,\chi(s_t),
\qquad
\bar\omega_T:=T^{-1}\sum_{t=1}^T\omega_t,
\end{equation}
where \(\hat\delta_t\) is the noisy TD residual of the main path.
The
corrected estimate is \(V^{\rm c}_T(s)=V(s;\bar\theta_T)+\chi(s)^\top\bar\omega_T\).
Since the recursion \eqref{eq:second_stage} depends on the evolving first stage parameter, we analyze the two estimators jointly.  Define \(e_t^*=\delta_t^*+\gamma\chi(s_t')^\top\omega^*-\chi(s_t)^\top\omega^*\),
\(B^*=\EE[\chi(s_t)\{\gamma\nabla_\theta V(s_t';\theta^*)
-\nabla_\theta V(s_t;\theta^*)\}^\top]\), 
\[
\mathcal J
=
\begin{pmatrix}
J^* & 0\\
B^* & -A_\chi
\end{pmatrix},
\quad\text{and }
\mathcal Q
=
\begin{pmatrix}
Q & \EE[\delta_t^*e_t^*\,\nabla_\theta V(s_t;\theta^*)\chi(s_t)^\top]\\
\EE[\delta_t^*e_t^*\,\chi(s_t)\nabla_\theta V(s_t;\theta^*)^\top]
& \EE[(e_t^*)^2\chi(s_t)\chi(s_t)^\top]
\end{pmatrix}.
\]

\begin{theorem}[Joint limit of calibration and correction]
\label{thm:joint_correction}
In the fast-decaying regime of Theorem~\ref{thm:clt_unified}, suppose \(\Sigma_\chi\succ0\) and, for the \(\kappa>0\) of Assumption~\ref{as:clt_common_poly}, \(\EE\|\chi(s_t)\|^{4+2\kappa}<\infty\) and \(\sup_{\|\theta-\theta^*\|\le r_0}\EE[|\delta(\theta;s_t,r_t,s_t')|^{2+\kappa} \|\chi(s_t)\|^{2+\kappa}]<\infty\). Then \((\theta_t,\omega_t)\to (\theta^*,\omega^*)\) a.s., and
\[
\sqrt T
\begin{pmatrix}
\bar\theta_T-\theta^*\\
\bar\omega_T-\omega^*
\end{pmatrix}
\Longrightarrow
\mathcal N\bigl(0,\;
(-\mathcal J)^{-1}\mathcal Q(-\mathcal J)^{-\top}\bigr).
\]
\end{theorem}

The Jacobian is block lower-triangular, so no stability condition beyond
Assumption~\ref{ass:regularity}(1) and \(\Sigma_\chi\succ0\) is needed. The
block \(B^*\) captures the estimation error that the TD residuals inherit from the first-stage calibration. By the delta method
with gradient \(h_s:=(\nabla_\theta V(s;\theta^*)^\top,\chi(s)^\top)^\top\),
\(\sqrt T\{V^{\rm c}_T(s)-V^{\rm c}(s)\}\Rightarrow
\mathcal N(0,h_s^\top(-\mathcal J)^{-1}\mathcal Q(-\mathcal J)^{-\top}h_s)\),
and by Proposition~\ref{prop:discrepancy_projection} this is a limit
theorem for the policy value when \(\Delta\) lies in the span of \(\chi\).

Running \eqref{eq:second_stage} inside each bootstrap path of
Algorithm~\ref{alg:bootstrap}, with the path's multiplier applied to the
second-stage increment, extends Theorem~\ref{thm:bootstrap_joint_limit} to
the stacked recursion and gives the corresponding \(t_{B-1}\) interval for
\(V^{\rm c}(s)\), as shown at the end of Appendix. 
For DF-Ind and DF-Fwd the second stage has no finite-difference bias. For
DF-Sym the midpoint residual adds a bias of order \(\nu_t^2\), which the
fast-decaying conditions  of Table~\ref{tab:query_noise_phase_diagram} make negligible.

\section{Discussion}
\label{sec:discussion}

The role of simulator specification extends beyond the definition of the calibration target. It also determines whether finite-difference approximation alters the population estimating equation and whether multiplier inference remains valid under Markov sampling. Under correct specification, the Bellman discrepancy vanishes at the calibration target, which preserves the target for DF-Ind and DF-Fwd even with a fixed perturbation radius and restores bootstrap validity along trajectories. Under misspecification, these cancellations generally fail, which changes both the attainable derivative-free rates and the validity of multiplier inference.

Several extensions remain open. Along Markov trajectories with a misspecified simulator, the fixed-\(B\) analysis could be extended to block multipliers or alternative random-scaling inference. Richer estimating equations, including multi-step residuals and more efficient instruments, may move the calibration target closer to the \(L_2\) projection while retaining the online structure. Adaptive replication schemes that allocate simulator rollouts according to observed variability are another natural direction.

\setlength{\bibsep}{0pt}
\bibliography{references}
\end{document}